\documentclass[onecolumn]{cohere}
\usepackage{lipsum}
\usepackage{stfloats} 
\usepackage{hyperref}
\usepackage{url}
\usepackage{booktabs}
\usepackage{graphicx}
\usepackage{multirow}
\usepackage[export]{adjustbox}
\usepackage{float}
\usepackage{caption}
\usepackage{subcaption}
\usepackage[colorinlistoftodos]{todonotes}
\usepackage{lscape}
\usepackage{rotating}
\usepackage{enumitem}
\usepackage{wrapfig}

\usepackage[symbol]{footmisc}

\title{Intriguing Properties of Quantization at Scale}

\author{
    name={Arash Ahmadian\textsuperscript{\textnormal{*\dag}}},
    affiliation={Cohere For AI },
}
\author{
    name={Saurabh Dash \textsuperscript{\textnormal{*}}},
    affiliation={Cohere},
}
\author{
    name={Hongyu Chen \textsuperscript{\textnormal{*}}},
    affiliation={Cohere},
}
\author{
    name={Bharat Venkitesh},
    affiliation={Cohere},
}
\author{
    name={Stephen Gou},
    affiliation={Cohere},
}
\author{
    name={Phil Blunsom},
    affiliation={Cohere},
}
\author{
    name={Ahmet Üstün},
    affiliation={Cohere For AI},
}
\author{
    name={Sara Hooker},
    affiliation={Cohere For AI},
}

\abstract{
Emergent properties have been widely adopted as a term to describe behavior not present in smaller models but observed in larger models \citep{wei2022emergent}. Recent work suggests that the trade-off incurred by quantization is also an emergent property, with sharp drops in performance in models over 6B parameters. In this work, we ask \textit{are quantization cliffs in performance solely a factor of scale?} Against a backdrop of increased research focus on why certain emergent properties surface at scale, this work provides a useful counter-example. We posit that it is possible to optimize for a quantization friendly training recipe that suppresses large activation magnitude outliers. Here, we find that outlier dimensions are not an inherent product of scale, but rather sensitive to the optimization conditions present during pre-training. This both opens up directions for more efficient quantization, and poses the question of whether other emergent properties are inherent or can be altered and conditioned by optimization and architecture design choices. We successfully quantize models ranging in size from 410M to 52B with minimal degradation in performance.}

\begin{document}

\section{Introduction}\label{sec:intro}
\begin{wrapfigure}{R}{0.5\linewidth}
    \includegraphics[width=1.0\linewidth]{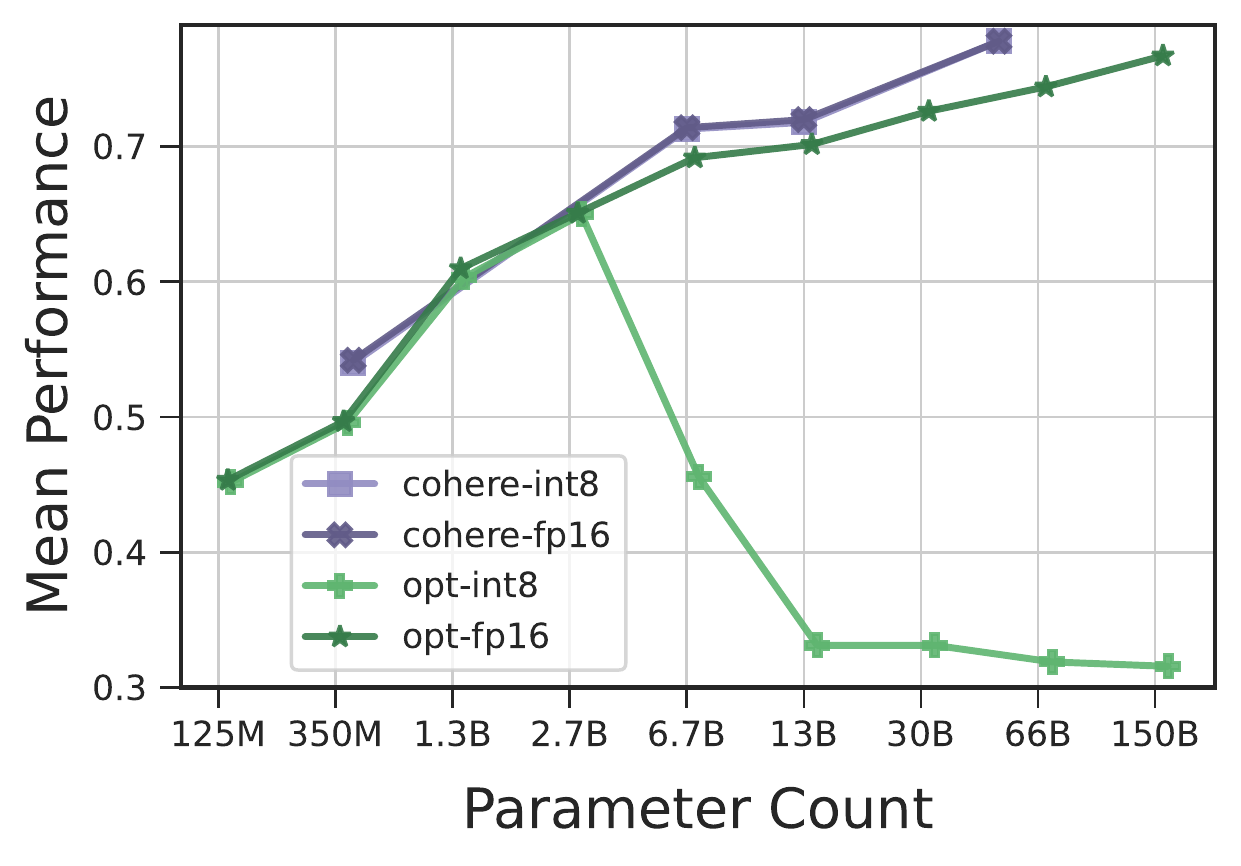}
    \caption{ Mean zero-shot accuracy on HellaSwag, PIQA, LAMBADA, and WinoGrad. In contrast to the OPT family, our models show minimal degradation after simple vectorwise quantization. Data points for OPT models are from \citep{dettmers2022}.}
    \label{fig:cohere_vs_OPT}
    \vspace{-0.5cm}
\end{wrapfigure}

\footnotetext[0]{\textsuperscript{*}Equal Contribution.}
\footnotetext[0]{\textsuperscript{\dag}Also affiliated with the University of Toronto \& the Vector Institute for Artificial Intelligence.}

The push for ever larger language models (LLMs) has been driven by a strong correlation between performance and the number of parameters \citep{chowdhery2022,OPT-zhang2022,Kaplan2020}. This has led to new breakthroughs in downstream performance, but has also posed new challenges in making these models accessible. Larger models incur higher memory and latency because of the requirement to store many more model weights and the optimizer in fixed memory \citep{dehghani2021,treviso2022}. Due to the massive size of state-of-art LLMs, inference often requires hosting across multiple machines which limits the practical usability of such models.

To address this, much research has focused on compression techniques such as quantization, which reduces the number of bits needed to represent each learned parameter in a model \citep{Gholami2021}. Quantization techniques are widely used in smaller model regimes -- quantizing weights stored as 32-bit or 16-bit floating-point numbers to 8-bit integers (\texttt{INT8}) produces large reductions in memory and latency.

However, at scale simple quantization techniques have been shown to lead to a pronounced degradation in performance \citep{Guangxuan2022}. This trade-off has been attributed by several recent works \citep{dettmers2022gptint,Zeng2022,Guangxuan2022,Bondarenko2021} to \textit{emergent outlier dimensions}---scaling transformer-based architecture results in large activation magnitude outliers which are concentrated along a few hidden dimensions. To remedy this degradation, mixed-precision solutions have been proposed that handle the outlier activations separately \citep{dettmers2022gptint}. While effective at preventing degradation, these specialized techniques pose significant latency overhead which negates some of the benefits to memory \citep{wu2020integer}. 

The difficulties of quantizing at scale prompt us to ask: \textit{are emergent properties due to nature or nurture?} Recent work introduces intriguing and somewhat contradictory answers to this question: Models like OPT-175B \citep{OPT-zhang2022} and FairSeq \citep{artetxe-etal-2022-efficient} 
exhibit pronounced sensitivity to post-training quantization and require complex mixed-precision decomposition quantization methods \citep{dettmers2022gptint,wei2022outlier,Bondarenko2021,Luo2020PositionalAP,Zeng2022}. 
On the contrary, BLOOM-176B \citep{scao2022bloom} is easier to quantize with a simple quantization recipe and a relatively small performance drop \citep{frantar2022, Guangxuan2022}. \citet{Zeng2022} hypothesize that the observed difference in weight distribution characteristics may be due to the difference in optimization choices made during pre-training. 

In this work, we seek to reconcile these observations. We posit that it is possible to optimize for a quantization friendly training recipe that suppresses large activation magnitude outliers. This leads to a distribution of activations and weights that are more amenable to simple INT8 quantization recipes and does not necessitate the need for complex and inefficient mixed-precision computations. Our results show that we can introduce simple INT8 post-training quantization with negligible impact on performance due to choices we make during the pre-training stage. As shown in Figure \ref{fig:cohere_vs_OPT}, across 8 zero-shot downstream tasks, our models do not present any significant performance drop, having only 0.24\% average degradation in a 52 billion parameter model. 

In summary, our contributions are as follows: 
\begin{itemize}
    \item We conduct a controlled large scale study -- we maintain the same architecture and vary key optimization choices such as weight decay, gradient clipping, dropout and precision of training representation. We present results across models varying from 410 million to 52 billion parameters, with each experiment variant trained from random initialization. While this requires a compute intensive set-up, it allows us to rigorously disentangle what factors actually influence sensitivity to quantization.
    \item We show that reoccurring activation outliers are not a universal \textit{emergent property} of LLMs at scale and can be avoided at scales as large as 52B given the right optimization choices. Our 52B parameter model shows only 0.26\% performance degradation across 8 tasks with INT8 PTQ quantization of \emph{both} activations and weights. 
    \item We contribute a fine-grained analysis of activations and weights, and show that several key weight and activation characteristics may explain the difference in sensitivity between our robust models and models like OPT which have been shown to have pronounced sensitivity to quantitization at scale. We hope these insights help guide future model design and pre-training strategies.
\end{itemize}

\section{Background}\label{sec:Preliminaries}

Quantization refers to compressing weights and activations of a neural network into lower-bit representations. Here, our focus is on  \textbf{one-shot post-training quantization} (PTQ) \citep{Guangxuan2022, dettmers2022gptint}, which quantizes the network post-training without additional finetuning steps for calibration. Given the complexities of successfully training a large language model \citep{OPT-zhang2022,rae2021}, PTQ methods are extremely attractive as these techniques require the least modification to pre-trained parameters, compared to quantization-aware training methods \citep{zafrir2019q8bert, krishnamoorthi2018quantizing} or quantization-aware finetuning \citep{park2022quadapter, yao2022zeroquant, frantar2022, zhuo2022empirical, BRECQ2021, hubara2020improving, nagel2020up} which both require updates to model weights. We include more detail about each of these broad groups of techniques in Appendix \ref{sec:ext-literature-review}.

To-date PTQ techniques that quantize both the activations and weights of a language model have proven extremely challenging at large scale ($>$6B parameters) leading to pronounced drops in performance. Instead, less aggressive techniques have been used such as \textit{weight-only quantization} \citep{llamacpp, frantar2022, Zeng2022} that leaves the activations in higher precision or \textit{quantization with mixed-precision decomposition} \citep{dettmers2022gptint} which decomposes the matrix multiplication to compute a small fraction of elements at a higher precision (FP16) while the bulk of the computations is performed at low precision (INT8). 

\textit{Weight-only quantization} brings speedup to inference by reducing the amount of data movement. However, as large language models are scaled, progressively they become compute-bound and the improvements due to weight-only quantization stagnate. While \textit{mixed-precision decomposition} approaches have theoretical latency benefits due to the bulk of the computation being performed at lower precision, in practice without specialized hardware \citep{9425549, 10.1145/3400302.3415679,Hooker2021}, GPU kernels, or additional kernel calls to prepare the inputs and weights for mixed-precision computation, the projected benefits cannot be realized \citep{dettmers2022gptint}. To further realize latency gains, we need to quantize both \textit{weights and activations} into 8-bit integers (INT8) to
utilize specialized integer INT8 GEMM kernels, which are supported by a wide range of hardware (e.g., NVIDIA GPUs, Intel CPUs, Qualcomm DSPs, etc.) In addition, weight and activations quantization further enables compressing key-value cache to INT8. Since key-value cache takes up a significant part of the GPU memory during inference \citep{sheng2023high}, weight and activations quantization further contributes to memory saving and high throughput inference.

Hence, the most challenging \textit{weights and activations} setting is the focus of this work. More concretely, given a neural network that learns a function \textit{f} parameterized by a set of weights $\{\mathbf{W_0},\mathbf{W_1},...,\mathbf{W_n}\}$ with corresponding activations $\{\mathbf{X_0},\mathbf{X_1},...,\mathbf{X_n}\}$, during quantization, activation and weight tensors denoted by $\mathbf{X} \in \mathbb{R}^{t \times h}$ and $\mathbf{W} \in \mathbb{R}^{h \times o}$ -- where $t$ denotes the sequence-length, $h$ the input hidden units and $o$ the output hidden units\textsuperscript{1}\footnotetext[0]{\textsuperscript{1}We omit the batch dimension for simplicity.} -- are replaced with lower-bit counterparts $\mathbf{W_Q}$ and $\mathbf{X_Q}$ by scaling and rounding the original tensors to INT8 range. We focus on \emph{vector-wise quantization} recipe \citep{dettmers2022gptint} to increase quantization granularity. In vector-wise quantization, we define the row-scaling vector $\mathbf{s_x} \in \mathbb{R}^{t}$ and column-scaling vector $\mathbf{s_w} \in \mathbb{R}^{o}$ by calculating the scaling constants for each row/column through uniform symmetric mapping \citep{nagel2021white}. After INT8 matrix multiplication, dequantization of $\mathbf{X_QW_Q}$ back to FP16 is through element-wise multiplication with $\mathbf{s_x}$ and $\mathbf{s_w}$:

\begin{equation}
\label{eq:vector-wise}
    \mathbf{XW} \approx \mathbf{s_{x}} \odot (\mathbf{X_Q W_Q}) \odot \mathbf{s_{w}}
\end{equation}

Where $\odot$ denotes broadcastable matrix multiplication. The quantization and dequantization steps for the above do not add much memory overhead compared to quantizing with a single scaling constant for each weight or activation tensor, while significantly increasing the representation power of INT8.

\section{Methodology and Experimental Setup}\label{sec:methodology}
\subsection{Methodology}

Our goal is to understand whether sensitivity to widely used quantization techniques is inherently an emergent property at scale or due to optimization choices made during pre-training. Recent work has presented seemingly contradictory empirical findings -- some models such as OPT-175B show pronounced sensitivity at scale to post-training quantization while other models such as BLOOM-176B are relatively robust to post-training quantization.
\begin{wraptable}{R}{0.44\linewidth}
\centering
\begin{tabular}{lr}
\toprule
Experimental Axes & Choices \\ \midrule
Weight decay & 0.001, ~0.01, ~0.1 \\
Gradient clipping &  None, ~~1 \\
Dropout & ~0,~0.1,~0.4, 0.8 \\
Half-precision & \texttt{bf16}, ~~\texttt{fp16}\\ 
\bottomrule
\end{tabular}
\caption{Optimization choices that are explored for pre-training in our controlled setup.}
\label{tab:experimental-axes}
\vspace{-0.5cm}
\end{wraptable}

These models differ in numerous ways such as architectural differences, pre-training data, training infrastructure, and finally optimization choices,making it challenging to attribute differences in quantization performance. To rigorously isolate what choices result in sensitivity to quantization, we measure the impact of optimization choices within a tightly controlled experimental setup -- training the same large scale model architecture from random initialization while rigorously varying only key aspects of the optimization procedure. Each optimization choice is evaluated in two ways: we measure the resulting degradation after PTQ in zero-shot downstream performance and then analyze the model weights and feature activations to understand how the characteristics at scale impact quantization performance.

Training multiple multi-billion parameter size language models is extremely expensive -- a single 52B language model  takes roughly 20 days of training with 2048 TPU cores.\textsuperscript{2}\footnotetext[0]{\textsuperscript{2}We include more details about the hardware and training requirements in Section \ref{sec:exp-setup}}. Therefore, we first conduct our controlled experiments on 410M and 6B models using early checkpoints and then validate the results at scale, by fully training 6B, 13B, and 52B parameter size models with our most quantization friendly training recipe. In practice, we found performance at early checkpoints predictive of fully trained model performance.

We briefly describe each of the axes of variations below:

\textbf{Weight decay} Weight decay is widely used to impede over-fitting by penalizing large magnitude weights \citep{goodfellow2016deep}. We experiment with a range of weight decay values \{\texttt{0.001}, \texttt{0.01}, \texttt{0.1}\}.

\textbf{Gradient clipping} Gradient clipping rescales the norm of the gradient vector if it exceeds the threshold \citep{pmlr-v28-pascanu13}. It is widely used in LLMs to prevent exploding gradients and accelerate training convergence \citep{GLM-Du2021,OPT-zhang2022}. We experiment with a gradient norm threshold of \texttt{1} as well as training without gradient clipping.

\textbf{Dropout} Dropout is a widely used regularization technique that drops neurons with a probability of $p$ during training \citep{dropout-journal}\citep{hinton2012dropout}. We apply dropout to the output of the self-attention block and the feed-forward block before the corresponding residual connection as described in \citet{vaswani2017attention}, but we do not use a dropout for the input embeddings. We experiment with \{\texttt{0}, \texttt{0.1}, \texttt{0.4}, \texttt{0.8}\} dropout probabilities.

\textbf{Half-precision data type: bf16 vs fp16} 
\label{sec:mixed-precision-training} Training neural networks in mixed-precision is a common technique to reduce the memory requirement and improves training time while often achieving comparable performance to full-precision training \citep{micikevicius2017mixed}. In this technique, a copy of weights is stored in full-precision (\texttt{fp32}) whereas the forward and backward passes are done using half-precision in either float16 (\texttt{fp16}) or bfloat16 (\texttt{bf16}) \citep{kalamkar2019bfloat,deanbf16,tensorflow2015}. 

We experiment with \texttt{fp16} and \texttt{bf16}. Furthermore, for each half-precision data type, we vary weight decay values of (\texttt{0.1}, \texttt{0.01}) to observe whether the effect of the half-precision data type is exasperated with a smaller weight decay value of \texttt{0.01}.

\begin{figure}[t]
    \centering
    \begin{subfigure}[t]{0.338\linewidth}
        \includegraphics[width=1.00\linewidth]{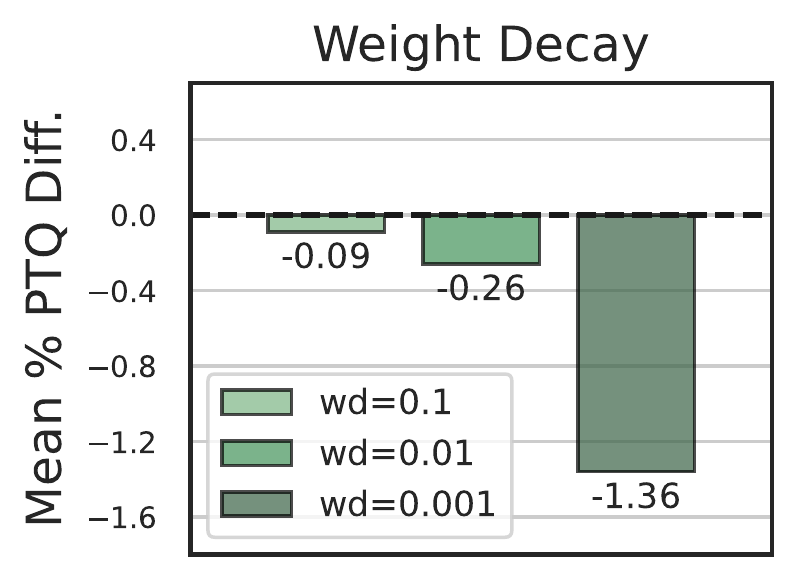}
        \caption{}
        \label{fig:wd}  
    \end{subfigure}
    \begin{subfigure}[t]{0.31\linewidth}
        \includegraphics[width=1.0\linewidth]{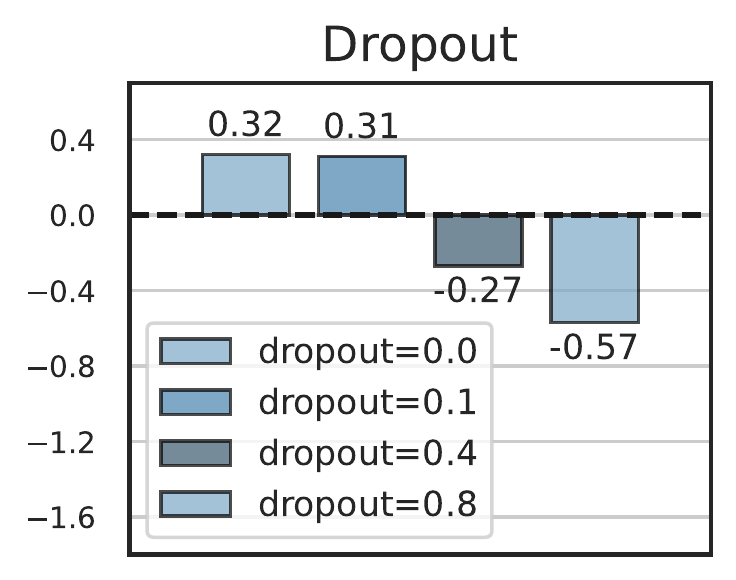}
        \caption{}
        \label{fig:dropout}  
    \end{subfigure}
    \begin{subfigure}[t]{0.31\linewidth}
        \includegraphics[width=1.0\linewidth]{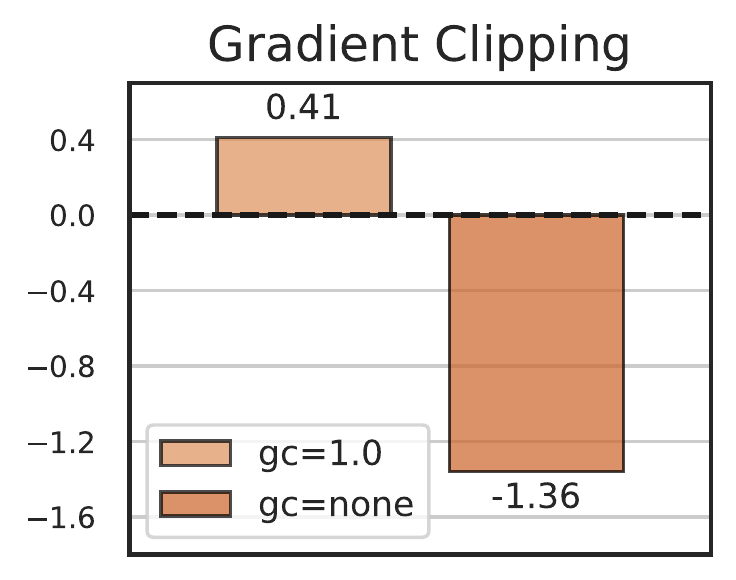}
        \caption{}
        \label{fig:gc}  
    \end{subfigure}
    \caption{Study of the PTQ performance when varying weight decay, dropout, and gradient clipping. In Figure \ref{fig:gc} a control weight decay value of 0.001 is used to to minimize the effects of weight decay when studying gradient clipping. Otherwise, we use weight decay of 0.1, dropout of 0, gradient clipping threshold of 1.0, and \texttt{bf16} training,  as control variables.}
\end{figure}

\subsection{Experimental Setup} \label{sec:exp-setup}

\textbf{Model} We train autoregressive decoder-only Transformer models \citep{Liu2018GeneratingWB} with a standard language modeling objective. Given an input sequence of $S=$ $\left[s_1, \cdots, s_t\right]$, a language model with parameters $\theta$ trained to minimizes the following negative log likelihood:  
\begin{gather}
    L(S)=\sum_{i}-log P(s_i|s_{<i}; \theta )
\end{gather}
Our language models follow the traditional GPT style architecture reported in \citet{radfordlanguage}. Different from the \citet{radfordlanguage}, we do not share the same weight matrix for input and output embedding layers. Instead, we learn separate input and output projections to enable higher model parallelism. We train models with parameter sizes ranging from 410 million to 52 billion. All models have a maximum sequence length of 2048 tokens. We use SentencePiece \citep{kudo-richardson-2018-sentencepiece} tokenizer with a vocabulary of 51200 to tokenize the text. 

\textbf{Training details} We pre-train models using a mixture of datasets from Common Crawl and C4 \citep{raffel2020exploring} with AdamW \citep{loshchilov2018decoupled} optimizer and a batch size of 256.
We use a cosine learning rate scheduler with 1500 warm-up steps. We use GeLU activations \citep{Hendrycks2016}. All the models are trained with mixed-precision, i.e. forward and backward passes are computed in \texttt{bf16} or \texttt{fp16} half-precision format, but the model parameters are stored in \texttt{fp32} in the distributed optimizer state \citep{10.5555/3433701.3433727}. For half-precision \texttt{fp16}, we experimented with a variant by switching layernorm arithmetic from \texttt{fp16} to \texttt{fp32} for better numeric stability \citep{micikevicius2017mixed}. Without layernorm arithmetic being in \texttt{fp32}, the training was unable to converge.

To avoid an exorbitant computational cost given each variant requires training from scratch and we are evaluating very large scale models, we first iterated on 410M models but observed very minimal degradation. As a result, we scaled to 6B parameters which is the scale at which we present our results. Following the analysis, we validate our findings by scaling the most PTQ friendly optimization choices to 13B and 52B models until convergence. The optimal training hyper-parameters for PTQ are provided in Appendix \ref{apendix:model_arch}. 

\textbf{Infrastructure}
We use TPU-v4 chips \citep{10.1145/3140659.3080246} to train, and Nvidia A100 GPUs to evaluate our models. All models are trained using the \texttt{FAX} \citep{Yoo2022} framework which enables efficient model and data parallelism. It takes approximately 72 hours on 128 cores to train a 6B parameter model for 75000 steps.

\textbf{Evaluation}  We evaluate each model variant on Copa (test and dev set) \citep{SuperGLUE}, \mbox{HellaSwag} \citep{HellaSwagCAZellers2019}, PIQAValidation \citep{PIQABisk2020}, StoryCloze \citep{storycloze-mostafazadeh-etal-2016-corpus}, WinoGrande \citep{sakaguchi2019winogrande}, Paralex \citep{Fader2013ParaphraseDrivenLF-paralex}, and LAMBADA \citep{paperno-etal-2016-lambada}. All evaluations were done in a zero-shot setting. In total, we benchmark on $8$ tasks comprised of Multiple choice (MC) completion, MC Co-referencing, Generation, and Question Answering (QA) types. Details of our evaluation suite are given in Appendix~\ref{apendix:eval_suite}. For each experimental variant, we report the average percent performance difference and the pre-quantization and post-quantization performances across tasks. The percent degradation is calculated by normalizing each task's absolute degradation by the corresponding pre-quantization performance. 

\begin{figure}[t]
\centering
    \includegraphics[width=.85\linewidth]{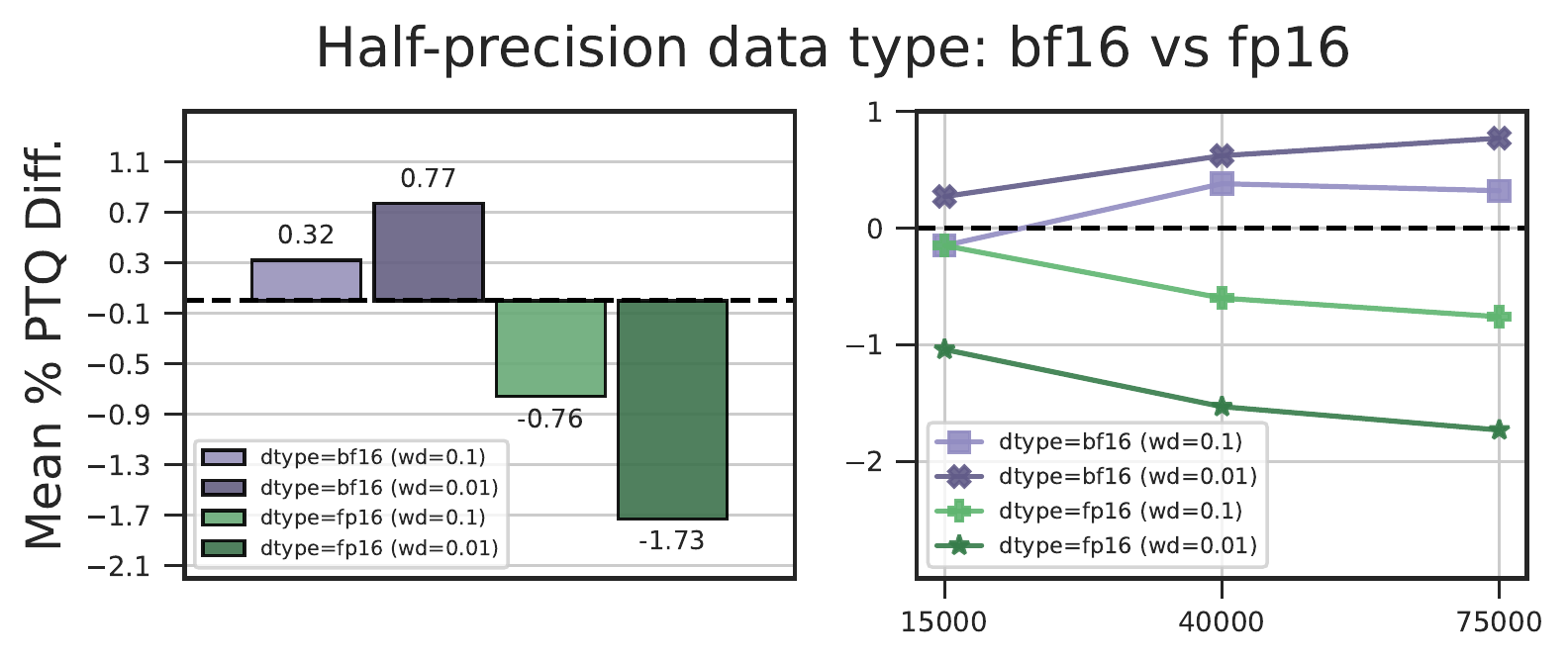}
    \caption{Study of PTQ performance when varying the precision used during training. On the left, the performance difference is plotted at 75000 steps whereas on the right, it is plotted over time. We observe that \texttt{fp16} training consistently leads to models which are far more sensitive to post-training quantization.}

    \label{fig:fp16_all}
\end{figure}

\section{Results and Discussion} \label{sec:results}

For each experimental axis, we train the corresponding variants to a maximum of 75000 steps. Below we present a breakdown of the degradation results and analysis for each experimental axis. All variants with the exception of \texttt{dropout=0.8}, had similar pre-quantization performance. This is important, as we are interested in comparing optimization choices that still result in models of comparable quality, but differing sensitivities to post-training quantization. Refer to Appendix \ref{apendix:result_breakdown} for the per-task breakdown of results.  

\textbf{Weight Decay} As can be seen in Figure \ref{fig:wd}, we observe that a higher level of weight decay during pre-training improves post-training quantization performance. We do not use gradient clipping in these experiments to isolate the impact of weight decay. A larger weight decay value (\texttt{0.1} vs \texttt{0.001}) results in better post-training performance (0.09\% vs 1.36\% degradation). Furthermore, as shown in Figure \ref{fig:fp16_all}, combining lower weight decay with \texttt{fp16} can further amplify sensitivity to post-training quantization. A small weight decay value (0.01) can cause higher performance degradation in post-quantization after training (1.73\%). 

\textbf{Dropout and Gradient Clipping} In Figure \ref{fig:dropout} we observe that higher levels of dropout correspond to sharper degradation in post-training quantization. Note that $P_{dropout}=0.8$ unsurprisingly leads to a poor absolute downstream performance, however, it helps establish a clear trend given other data points. Figure \ref{fig:gc} shows the relative quantization degradation for models with and without gradient clipping. When varying gradient clipping, a control weight decay value of 0.001 is used to minimize the impact of weight decay. As seen in the figure, gradient clipping shows a positive impact on the quantization performance, improving robustness to post-training quantization. This suggests that gradient clipping to an extent counteracts the effects of a small weight-decay value which would otherwise lead to higher quantization degradation. 

\textbf{Half-precision: bf16 vs fp16} Figure \ref{fig:fp16_all} shows the quantization degradation and absolute performance for \texttt{fp16} and \texttt{bf16} for 6B parameter models. Training with \texttt{fp16} leads to higher quantization degradation than \texttt{bf16}. We relate the degradation results to the numerical stability in training. \texttt{fp16} format uses a smaller range for the exponent than \texttt{bf16}. While \texttt{bf16} uses 8 bits for the exponent, \texttt{fp16} only uses 5. Most floating point formats also have denormalized numbers which allow for a soft underflow. This can get exponentially closer to 0.0f for each additional bit in the mantissa. This makes underflow more of a concern for floating point formats.

Notably, we observe that our findings provide insights into the quantization robustness of BLOOM-176B, which to our knowledge is the only open-source LLM (with more than 50B parameters) and a decoder-only block architecture, that is trained with \texttt{bf16}; compared to simliar sized \texttt{fp16} trained models such as OPT-175B \citep{Guangxuan2022}.

\textbf{Using Early Checkpoints to Infer Converged Model Behavior} Given the considerable computational cost of our experiments, it is valuable to explore whether converged model behavior can be inferred from checkpoints from early snapshots of training. In Figure \ref{fig:fp16_all}, we plot the relative post-training quantization degradation given checkpoints trained with different levels of precision at different steps during pre-training. We observe that quantization degradation increases with step count, but the relative impact of varying the bit representation emerges early in training which confirms the main trend. Interestingly, \texttt{fp16 (wd=0.01)} variant exhibits high quantization degradation in the starting phase training as early as 15000 steps.

\textbf{Scaling Insights to 52B scale} To validate our experimental findings at scale and with fully trained models, we pre-train 410M, 6B, 13B, and 52B parameter models using the best optimization choices with respect to robustness in post-training quantization: weight decay of 0.1, no dropout, gradient clipping of 1, and \texttt{bf16} as the half-precision format. Figure \ref{fig:cohere_vs_OPT} shows mean zero-shot accuracy for the non-quantized and quantized model using INT8 weights and activations. Compared with OPT models \citep{OPT-zhang2022}, our fully-trained models are significantly more robust to post-training quantization starting from 6B parameter size. Our largest scale model with 52B parameters, shows a  \textbf{0.08\%} \emph{improvement} in average performance across the evaluation suite and only \textbf{0.01\%} degradation across LAMBADA, HellaSwag, and PIQA where OPT-66B which is the closest OPT model in terms of size, has an extreme drop of $\sim$42\% as reported in \cite{dettmers2022gptint}.

To evaluate our models' performance under a different quantization recipe in addition to INT8, we also test 4-bit integer (INT4) column-wise weight-only quantization, similar to \cite{GLM-Du2021}. Our 52B parameter model exhibited only a 3.6\% relative drop in mean zero-shot performance across the 8 evaluation tasks. It is worth noting that this quantization scheme does not require any fine-tuning or optimization and hence these results highlight the high robustness of our trained models. In comparison, when applying the same quantization scheme to BLOOM, BLOOM-176B and BLOOM-7B show 29.5\% and 18.7\% degradation respectively on LAMBADA \citep{GLM-Du2021}, while our 52B model only has 8.6\% degradation.

\iffalse
\begin{figure}[!th]
\centering
    \includegraphics[width=.49\linewidth]{Figures/rmse_plots/dtype-wd001.pdf}
    \includegraphics[width=.49\linewidth]{Figures/rmse_plots/dtype-wd01.pdf}
    \includegraphics[width=0.49\linewidth]{Figures/rmse_plots/wd.pdf}
    \includegraphics[width=0.49\linewidth]{Figures/rmse_plots/gc.pdf}
    
    \caption{Histogram of per-token RMSE measured from a unifrom distribution with the same range. Models with higher PTQ degradation, show heavier tails.}
    \label{fig:fp16_all}
\end{figure}
\fi

\section{Weight and Activation Analysis} \label{sec:weights_acts}

Our results in Section \ref{sec:results} find that sensitivity to quantization at scale is not an inherent emergent property. Rather, it can be avoided at scales as large as 52B given the right optimization choices. In this section, we perform a fine-grained analysis of activations and weights to understand how the trained distribution of our models differs from models like OPT that are far more sensitive to quantization. For all the metrics proposed below, we include the complete analysis for all layers in Appendix \ref{apendix:extended_analysis}.

\begin{figure}[t]
  \centering\includegraphics[width=0.48\linewidth]{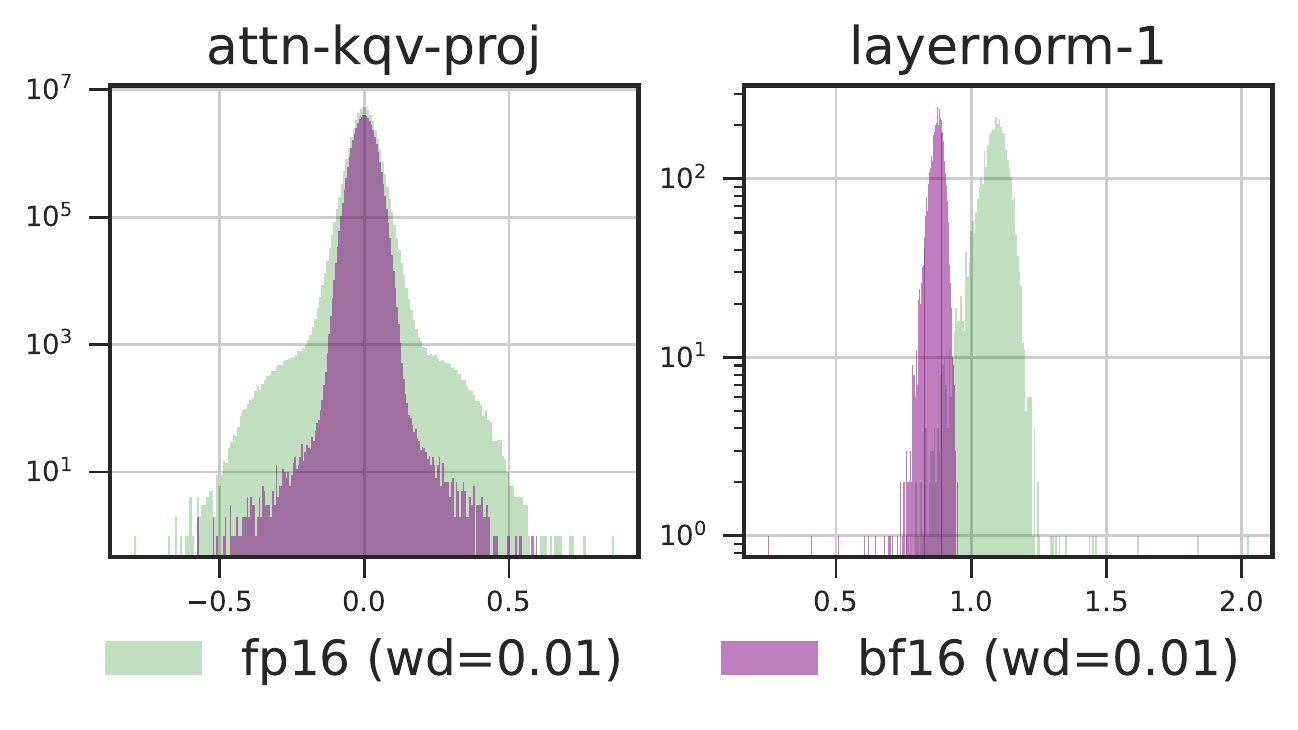}
  \centering\includegraphics[width=0.487\linewidth]{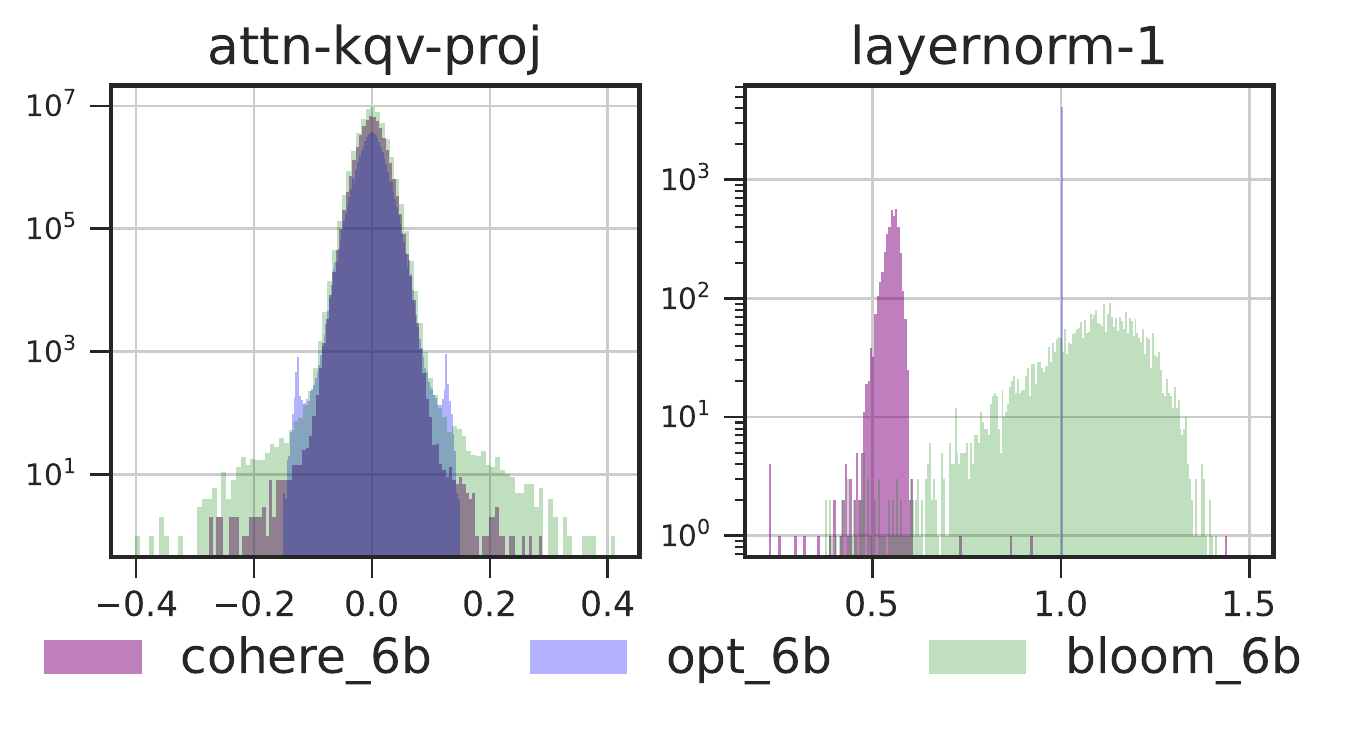}
  \caption{Weight distribution of \texttt{attn-kqv-proj} and \texttt{layernorm} gain ($\mathbf{g}$) parameter in an example block (Block 14) for both \texttt{fp16}/\texttt{bf16} variants and our final 6B model in comparison with OPT and BLOOM. Weight distributions for all blocks are shown in Appendix~\ref{sec:all-weigh-distributions}}
\label{fig:weights_and_ln}
\end{figure}

\begin{figure}[t]
    \centering
    \includegraphics[width=1.0\linewidth]{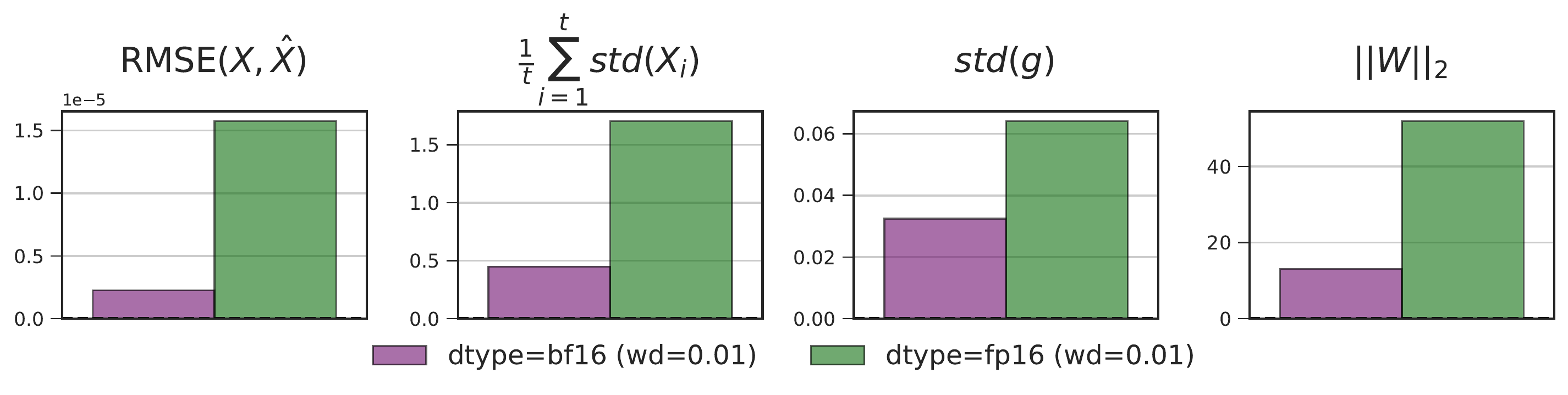}
    \includegraphics[width=1.0\linewidth]{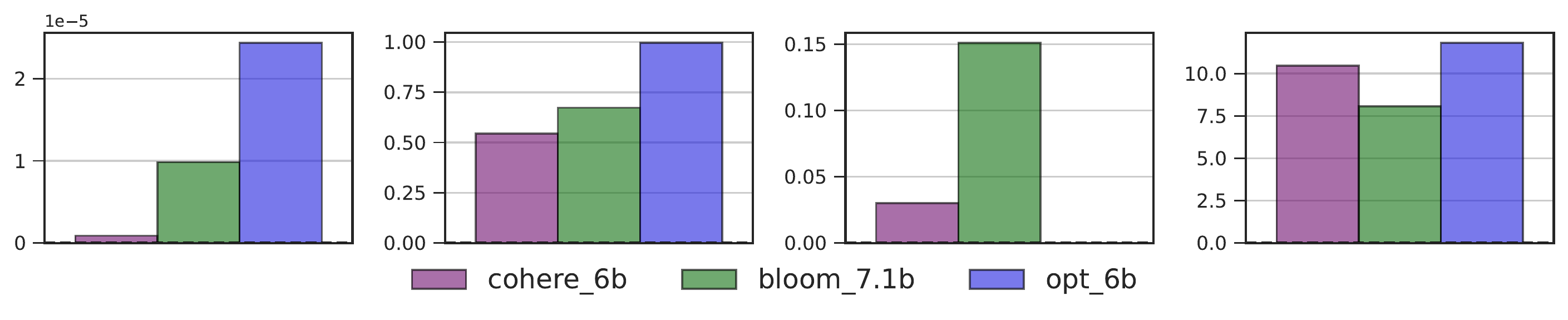}
    \caption{Average input token activation STD, preceding layernorm gain STD, and spectral norm of linear weights, follow trends similar to those observed in RMSE($\mathbf{X}$,$\mathbf{\hat{X}}$). Plots correspond to the \texttt{attn-kqv-proj} layer in an example block (Block 14). Comparisons for all other blocks are given in Appendix \ref{apendix:all-act-distributions}. \textbf{Top row:} Comparison of \texttt{fp16} and \texttt{bf16} variants. \textbf{Bottom row:} Our 6B model trained with optimal hyper-parameters compared against similar sized OPT-6B and BLOOM-7.1B models}
    \label{fig:fp16_layer13_stats}
    
\end{figure}

\textbf{Activations} As a first step, we analyze input activations of the attention projection \linebreak(\texttt{attn-kqv-proj}) as it is the earliest point for INT8 multiplication in a decoder block. Here, we measure root-mean-square error $\text{RMSE}(\mathbf{X},\mathbf{\hat{X}})$ where $\mathbf{\hat{X}}$ denotes the de-quantized activations. Additionally, we report the mean standard deviation of the input activations measured per token. While $\text{RMSE}(\mathbf{X},\mathbf{\hat{X}})$ directly indicates the quantization error, standard deviation (STD) has been shown to be closely related to the expected quantization error of a normally distributed input \citep{kuzmin2022fp8}. Figure \ref{fig:fp16_layer13_stats} compares the \texttt{bf16} and \texttt{fp16} variants. We observe that the RMSE and STD of \texttt{fp16} are far higher than the \texttt{bf16} variant -- the RMSE for the \texttt{fp16} variant is 6.9x the RMSE for the \texttt{bf16} variant. This difference is even more pronounced if we compare our model to the OPT: the RMSE and STD of the OPT are 27.7x and 1.8x higher respectively relative to our model (Figure \ref{fig:fp16_layer13_stats}; Bottom row).  

\textbf{LayerNorm} Since we use a \emph{pre-norm} architecture, the input activations to \texttt{attn-kqv-proj} and \texttt{outer-expander-mlp} are both preceded by a layernorm. The layernorm gain parameter, $\mathbf{g} \in \mathbb{R}^h$, directly influences the output activation tokens' spread, and can significantly vary in distribution shape as seen in Figure \ref{fig:weights_and_ln}. We generally observe that within our experimental axes, the standard deviation of the gain parameters for both the first and second layernorms are higher in a significant number of layers for variants with higher degradation compared to others in the same axis. In Figure \ref{fig:fp16_layer13_stats}, we compare the standard deviation of $\mathbf{g}$ for self-attention layernorm and we observe that STD($\mathbf{g}$) is 2x higher for the \texttt{fp16} variant relative to \texttt{bf16}. Our findings add further support to previous work which suggests that the gain parameters act as an outlier amplifier and thus further quantization degradation through scaling \citep{wei2022outlier}.

In the bottom row of Figure \ref{fig:fp16_layer13_stats}, we also compare STD($\mathbf{g}$) of our model relative to OPT and BLOOM. We observe that even the BLOOM model that is relatively robust to quantization has a far larger STD($\mathbf{g}$) than our model with a multiplier of 5x. Interestingly, we find that OPT-6B layernorm gain parameters are all set to 1.0 while biases varied as expected. Hence, given the gain parameters appear to be hardcoded, the STD($\mathbf{g}$) of the OPT model is 0. We were not able to find any mention of such design decision either in \cite{OPT-zhang2022} or the github repository: \url{https://github.com/facebookresearch/metaseq}. 

\textbf{Attention Projection Weights} Finally, we compare the weight distribution of \texttt{attn-kqv-proj} layers. As seen in Figure \ref{fig:weights_and_ln},  the \texttt{fp16} variant has a significantly wider distribution compared to \texttt{bf16}.  
Additionally, inspired by \cite{lin2019defensive}, we use spectral norm to measure the maximum degree of noise amplification for each token activation.

For a given weight matrix $\mathbf{W} \in \mathbb{R} ^ {h \times o}$, and input token activation noise $\mathbf{x_{\delta} \in \mathbb{R} ^ {h}}$, the spectral norm $\|\cdot\|_2$ is defined as 
\begin{equation}
\label{eq:singular}
    \|\mathbf{W}\|_{2} = \sup_{\mathbf{x_\delta}\neq0}\frac{{\| \mathbf{W{x_\delta}}\|}_2}{\|\mathbf{x_\delta}\|_2} 
    = \sigma_{max}
\end{equation}

where $\sigma_{max}$ is the largest singular value of $\mathbf{W}$.  

As seen in Figure \ref{fig:fp16_layer13_stats}, we observe that the spectral norm of the \texttt{fp16} variant is 4x higher than the \texttt{bf16}. 
In addition, on the bottom row of Figure \ref{fig:fp16_layer13_stats}, we observe that both BLOOM and our model have generally lower spectral norm than OPT 6B that is far more sensitive to quantization.

\textbf{Discussion} In addition to the metrics proposed above, we also sought to incorporate recent work that proposes to use the number of \emph{outlier dimensions} as a proxy measure to understand sensitivity to quantization degradation \citep{dettmers2022gptint}. Activation feature dimensions are classified as outlier dimensions 
when the activation values (denoted as $\alpha$) are greater than 6.0 at more than 25\% of layers and 6\%  of tokens. However, we find that a threshold of 6.0 is too high to classify a feature dimension as an outlier for all the variants we consider. After correspondence with the authors, we also explored various adaptations of this outlier detection recipe presented in \cite{dettmers2022gptint} to make it generalizable. However, we did not observe a clear correlation between these measures and sensitivity to quantization. We refer to Appendix \ref{apendix:outlier} for detailed treatment of these replication efforts. Due to these observations, we hope that the metrics we proposed and evaluated in this Section will help further discussion about useful proxy metrics for guiding  pre-training optimization choices to improve robustness to quantization.

\section{Related Work} \label{sec:related}

\textbf{Challenges of Quantization at Scale} Recently, there have been several studies to characterize the emergence of outliers at scale, and relate this to the difficulties in post-training quantization of both weights and activations \citep{dettmers2022gptint, wei2022outlier, puccetti2022}. \cite{dettmers2022gptint} depict a phenomenon of \textit{emerging outliers} by observing that large outlier dimensions systematically emerge at a certain scale (6.7B parameters) which hamper quantization attempts. Extreme outliers at scale was also empirically confirmed in follow-up works \citep{Zeng2022, Guangxuan2022}.
The causes of outliers have also been the subject of recent work. \citet{puccetti2022} observe that in Masked Language Models (MLMs) the magnitude of hidden state coefficients corresponding to outlier dimensions correlates with the frequency of encoded tokens in pre-training data. \citet{wei2022outlier} observe that LayerNorm scaling ($\mathbf{g}$) amplifies the outliers and can be suppressed using a modified LayerNorm and token-wise clipping. \citet{wortsman2023stable} consider large-scale vision-language models and show that quantization techniques are more stable if the network is trained and initialized so that large feature magnitudes are discouraged. 

Most mitigation strategies to quantize in the presence of outliers has required more complex quantization techniques. For example,  \citet{dettmers2022gptint} propose selective mixed-precision computation by only computing the outliers at higher precision. However, such a setup proves difficult to map to hardware, limiting the inference speedup. 
\citet{Guangxuan2022} propose to smoothen out these outliers by migrating some of the activation variances into the model weights with appropriate scaling. Although the authors demonstrate the ability of this framework to scale to large models, additional rescaling is required for activations which leads to additional latency overhead without specialized kernels. Another limitation of \citet{Guangxuan2022} is that it relies on the assumption that outliers exist in activations, and that weights can bear additional outliers and still be easy to quantize. 

Our work is the first to show that outliers are not inherent to scaling large language models. Rather than an emerging property, they are a result of particular training methods. Compared to previous methods using extensive quantization schemes with custom kernels \citep{dettmers2022gptint}, our work applies PTQ using simple, one-shot linear weight and activation quantizations which can take advantage of NVIDIA-provided CUTLASS kernels, leading to a significant decrease in latency and memory footprint.

\section{Conclusion} \label{sec:conclusion}

We present a rigorous study of the effect of how various optimization choices affect INT8 PTQ with the goal of reconciling the recent contradictory observations regarding emergent properties in Large Language Models. We show that regularization directly impacts PTQ performance and that higher levels of regularization through common techniques such as weight-decay, and gradient-clipping leads to lower post-training quantization degradation. We further demonstrate that the choice of half-precision training data type has a significant impact on PTQ performance -- emergent features are significantly less pronounced when training with \texttt{bf16}.

\textbf{Broader Impact}  Our work serves as a useful counter-example to scholarship which has advanced the notion that certain properties depend only on model scale \citep{wei2022emergent}. Rather, our results support the conclusion that optimization choices play a large role in whether emergent properties are present. We believe there is more work to be done here. We also hope that the insights gained from our work illustrate the significant impact the underlying hardware can have on PTQ. Currently, \texttt{bf16} training is possible on TPUs and only very recently introduced to A100 \& H100 GPUs. Finally, we belive our results present an impactful formula for training models which are inherently easier to quantize at scale, making these models more accessible for deploying in a variety of deployment environments.

\textbf{Limitations} We do not vary the architectural design and training objective in our experiments given our goal of a controlled experimental set-up and the large computational cost of each variant. We leave exploring the impact of different training objectives and architecture design choices to future work.

\section{Acknowledgement}
We thank João Araújo, Milad Alizadeh and other colleagues in Cohere \& Cohere For AI for helpful feedback and support. We also thank Tim Dettmers for assisting in replicating the outlier dimension definition and results in  \texttt{int8.LLM()}. 

\newpage
\bibliography{references.bib}
\newpage

\appendix
\section*{Appendix}

\section{Extended Results \& Architecture Details} 
\label{apendix:extended_results}

\subsection{Optimal Hyper-parameters}\label{apendix:model_arch}

\begin{table*}[!ht]
\centering
\newcommand*{\cindent}{\hspace*{0.5cm}}
\newcommand*{\cindentt}{\hspace*{0.75cm}}

\begin{tabular}
{p{0.21\linewidth}p{0.21\linewidth}p{0.21\linewidth}p{0.24\linewidth}}
    \bottomrule
    \\[-0.8em]
    Weight decay & Gradient clipping & Dropout & Half-precision datatype\\
    \midrule
    \cindent0.1 & \cindent1.0 & \cindent0 & \cindentt\texttt{bf16} \\
    \bottomrule
\end{tabular}
\caption{Optimal hyper-parameters for PTQ based on results in Section \ref{sec:results}}

\label{tab:model_arch}
\end{table*}

\subsection{Evaluation Suite} \label{apendix:eval_suite}
Below is a detailed breakdown of the evaluation suite we evaluate our models with. 

\begin{table*}[ht]
\centering
\begin{tabular}[t]{lccc}
\toprule
Benchmark &Task Type & Evaluation Metric \\
\midrule
Copa (test set)\citep{SuperGLUE} & MC Completion & MC Accuracy \\
Copa100 (dev set) \citep{SuperGLUE}  & MC Completion & MC Accuracy  \\ 
HellaSwag  \citep{HellaSwagCAZellers2019} & MC Completion & MC Accuracy \\
PIQAValidation \citep{PIQABisk2020} & MC Completion & MC Accuracy  \\ 
StoryCloze \citep{storycloze-mostafazadeh-etal-2016-corpus} & MC Completion & MC Accuracy \\
WinoGrande \citep{sakaguchi2019winogrande} & MC Co-referencing & MC Accuracy \\
Paralex \citep{Fader2013ParaphraseDrivenLF-paralex} & Generation  & Likelihood (bytes) \\
LAMBADA \citep{paperno-etal-2016-lambada} & QA & Exact String Match Accuracy  \\
\bottomrule
\end{tabular}
\caption{An Overview of the 8 tasks we benchmark the zero-shot downstream performance of trained models. QA and MC denotes Question Answering, and Multiple-choice respectively.}
\label{tab:eval-suite}
\end{table*}
\newpage
\subsection{Task Result Breakdown}\label{apendix:result_breakdown}
\begin{table}[!htp]
\label{tab:prod_results}
\resizebox{\linewidth}{!}{\begin{tabular}{llcccccccccc}

\toprule
\textbf{Model Size} &  \textbf{Data type} &  \textbf{PIQA} &  \textbf{HellaSwag} &  \textbf{WinoGrande} &  \textbf{LAMBADA} &  \textbf{Copa} &  \textbf{Copa100} &  \textbf{StoryCloze} &  \textbf{Paralex} &  \textbf{Average}  \\
\midrule
       \multirow{3}{*}{52B} &            FP16 &                                   83.19 &                               82.48 &                               70.01 &                            75.47 &                        79.40 &                           81.00 &                               85.87 &                                   61.05 & 77.31 \\
        & W8A8 &                                   83.20 &                               82.40 &                               70.00 &                            75.50 &                        79.40 &                           82.00 &                               85.50 &                                   61.10 & \textbf{77.39} \\
        &   W4 &                                   80.20 &                               72.22 &                               66.30 &                            66.85 &                        78.20 &                           83.00 &                               82.56 &                                   60.43 & 73.72 \\ \midrule
       \multirow{3}{*}{13B} &            FP16 &                                   79.54 &                               75.26 &                               62.27 &                            70.81 &                        76.00 &                           76.00 &                               82.11 &                                   60.68 & 72.83 \\
        & W8A8 &                                   79.20 &                               74.60 &                               62.90 &                            69.90 &                        76.00 &                           75.00 &                               82.20 &                                   60.90 & \textbf{72.59} \\
        &   W4 &                                   76.66 &                               60.59 &                               57.30 &                            46.05 &                        73.60 &                           76.00 &                               77.40 &                                   59.74 & 65.92 \\ \midrule
        \multirow{3}{*}{6B} &            FP16 &                                   79.50 &                               74.20 &                               61.20 &                            70.50 &                        75.40 &                           79.00 &                               81.50 &                                   60.10 & 72.67 \\
        & W8A8 &                                   79.50 &                               73.70 &                               61.40 &                            70.00 &                        74.60 &                           77.00 &                               81.00 &                                   60.20 & \textbf{72.18} \\
        &   W4 &                                   76.93 &                               62.92 &                               56.43 &                            55.40 &                        74.00 &                           72.00 &                               77.21 &                                   59.01 & 66.74 \\ \midrule
      \multirow{3}{*}{410M} &            FP16 &                                   70.40 &                               46.90 &                               50.80 &                            48.80 &                        65.40 &                           65.00 &                               70.50 &                                   57.10 & 59.36 \\
        & W8A8 &                                   70.00 &                               46.80 &                               51.50 &                            47.80 &                        64.00 &                           64.00 &                               69.70 &                                   57.00 & \textbf{58.85} \\
        &   W4 &                                   67.19 &                               43.08 &                               50.59 &                            37.71 &                        62.80 &                           64.00 &                               67.47 &                                   54.60 & 55.93 \\
\bottomrule

\end{tabular}}
\vspace{0.1cm}

\caption{Our fully trained models with hyper-parameters outlined in Table \ref{tab:model_arch} show minimal PTQ degradation. }
\end{table}

\begin{adjustbox}{angle=90}
\centering
\resizebox{\textheight}{!}{
\centering
%\vspace{10cm}
\begin{tabular}{llcccccccccc}
\toprule
{} &  \textbf{Data type} &  \textbf{PIQA} &  \textbf{HellaSwag} &  \textbf{WinoGrande} &  \textbf{LAMBADA} &  \textbf{Copa} &  \textbf{Copa100} &  \textbf{StoryCloze} &  \textbf{Paralex} &  \textbf{Average}  &  \textbf{Average \% Diff}  \\
\midrule
\multirow{2}{*}{\texttt{wd=0.1 (gc=none)}}         &  
FP16 &           75.35 &60.58 &55.41 &56.59 &73.20 &71.00 &77.02 &58.75 &65.99 & \multirow{2}{*}{-0.09} \\[0.18cm]
& INT8 & 75.35 &60.36 &55.25 &57.69 &73.20 &70.00 &76.70 &58.62 &65.90 & \\[0.18cm]

\multirow{2}{*}{\texttt{wd=0.01 (gc=none)}}         &  FP16 &           75.03 &      60.55 &       55.25 &    59.01 & 72.00 &    69.00 &       77.53 &    58.99 &     65.92 & 
\multirow{2}{*}{-0.26} \\[0.18cm]
    &  INT8 &           74.48 &      60.10 &       55.49 &    60.14 & 72.40 &    67.00 &       77.21 &    58.87 &     65.71 &              \\[0.18cm]
\multirow{2}{*}{\texttt{wd=0.001 (gc=none)}}         &  FP16 &           75.90 &      60.71 &       55.80 &    58.16 & 73.00 &    71.00 &       76.64 &    58.50 &     66.21 &             \multirow{2}{*}{-1.36} \\[0.18cm]
   &  INT8 &           75.63 &      60.52 &       54.78 &    55.15 & 71.80 &    70.00 &       77.21 &    57.96 &     65.38 &             \\[0.07cm] \midrule
\multirow{2}{*}{\texttt{dtype=bf16 (wd=0.1)}}      &  FP16 &           75.68 &      60.92 &       55.96 &    57.60 & 71.40 &    71.00 &       75.81 &    59.05 &     65.93 &              \multirow{2}{*}{0.32} \\[0.18cm]
  &  INT8 &           75.95 &      60.70 &       56.83 &    58.32 & 71.40 &    71.00 &       75.62 &    59.06 &     66.11 &              \\[0.18cm] 
\multirow{2}{*}{\texttt{dtype=fp16 (wd=0.1)}}       &  FP16 &           73.83 &      59.96 &       55.96 &    56.14 & 72.00 &    67.00 &       75.94 &    58.33 &     64.89 &             \multirow{2}{*}{-0.76} \\[0.18cm]
  &  INT8 &           74.16 &      59.75 &       55.64 &    56.05 & 71.60 &    64.00 &       75.88 &    58.12 &     64.40 &             \\[0.07cm] \midrule
\multirow{2}{*}{\texttt{dtype=bf16 (wd=0.01)}}      &  FP16 &           74.97 &      60.79 &       55.49 &    57.52 & 72.00 &    68.00 &       75.88 &    58.74 &     65.42 &              \multirow{2}{*}{0.77} \\[0.18cm]
 &  INT8 &           75.08 &      60.51 &       55.88 &    59.31 & 72.60 &    69.00 &       76.00 &    58.84 &     65.90 &             \\[0.18cm]
\multirow{2}{*}{\texttt{dtype=fp16 (wd=0.01)}}      &  FP16 &           74.81 &      58.11 &       54.93 &    57.31 & 70.20 &    71.00 &       74.67 &    58.25 &     64.91 &             \multirow{2}{*}{-1.73} \\[0.18cm]
 &  INT8 &           73.61 &      56.90 &       54.22 &    53.02 & 71.60 &    69.00 &       74.67 &    57.92 &     63.87 &          \\[0.07cm] \midrule
\multirow{2}{*}{\texttt{gc=1.0 (wd=0.001)}}        &  FP16 &           74.65 &      60.03 &       54.78 &    59.01 & 71.60 &    67.00 &       76.96 &    58.92 &     65.37 &              \multirow{2}{*}{0.41} \\[0.18cm]
   &  INT8 &           74.92 &      59.91 &       54.62 &    59.69 & 72.20 &    68.00 &       76.96 &    58.90 &     65.65 &              \\[0.18cm]
\multirow{2}{*}{\texttt{gc=none (wd=0.001)}}      &  FP16 &           75.90 &      60.71 &       55.80 &    58.16 & 73.00 &    71.00 &       76.64 &    58.50 &     66.21 &             \multirow{2}{*}{-1.36} \\[0.18cm]
   &  INT8 &           75.63 &      60.52 &       54.78 &    55.15 & 71.80 &    70.00 &       77.21 &    57.96 &     65.38 &             \\[0.07cm] \midrule
\multirow{2}{*}{\texttt{dropout=0.0}}              &  FP16 &           75.68 &      60.92 &       55.96 &    57.60 & 71.40 &    71.00 &       75.81 &    59.05 &     65.93 &              \multirow{2}{*}{0.32} \\[0.18cm]
        &  INT8 &           75.95 &      60.70 &       56.83 &    58.32 & 71.40 &    71.00 &       75.62 &    59.06 &     66.11 &              \\[0.18cm]
\multirow{2}{*}{\texttt{dropout=0.1}}            &  FP16 &           74.76 &      58.87 &       54.38 &    57.23 & 71.60 &    68.00 &       76.45 &    58.36 &     64.96 &              \multirow{2}{*}{0.31} \\[0.18cm]
         &  INT8 &           74.27 &      58.70 &       54.85 &    58.35 & 71.80 &    68.00 &       76.96 &    58.18 &     65.14 &              \\[0.18cm]
\multirow{2}{*}{\texttt{dropout=0.4}}                  &  FP16 &           74.92 &      55.80 &       54.70 &    58.98 & 71.00 &    66.00 &       74.03 &    57.91 &     64.17 &             \multirow{2}{*}{-0.27} \\[0.18cm]
       &  INT8 &           74.76 &      55.77 &       54.14 &    59.69 & 69.40 &    66.00 &       74.09 &    57.95 &     63.98 &             \\[0.18cm]
\multirow{2}{*}{\texttt{dropout=0.8}}                  &  FP16 &           67.79 &      30.87 &       50.51 &    37.12 & 67.00 &    65.00 &       61.94 &    29.16 &     51.17 &             \multirow{2}{*}{-0.57} \\[0.18cm] 
          &  INT8 &           67.79 &      30.75 &       50.12 &    36.52 & 66.60 &    65.00 &       61.74 &    28.91 &     50.93 &             \\[0.18cm]
\bottomrule
\end{tabular}}
\end{adjustbox}

\section{Extended Weight \& Activation Analysis}\label{apendix:extended_analysis}

\subsection{Weight Distributions}
\label{sec:all-weigh-distributions}

\begin{figure}[!ht]
    \centering
    \includegraphics[height=0.88\textheight]{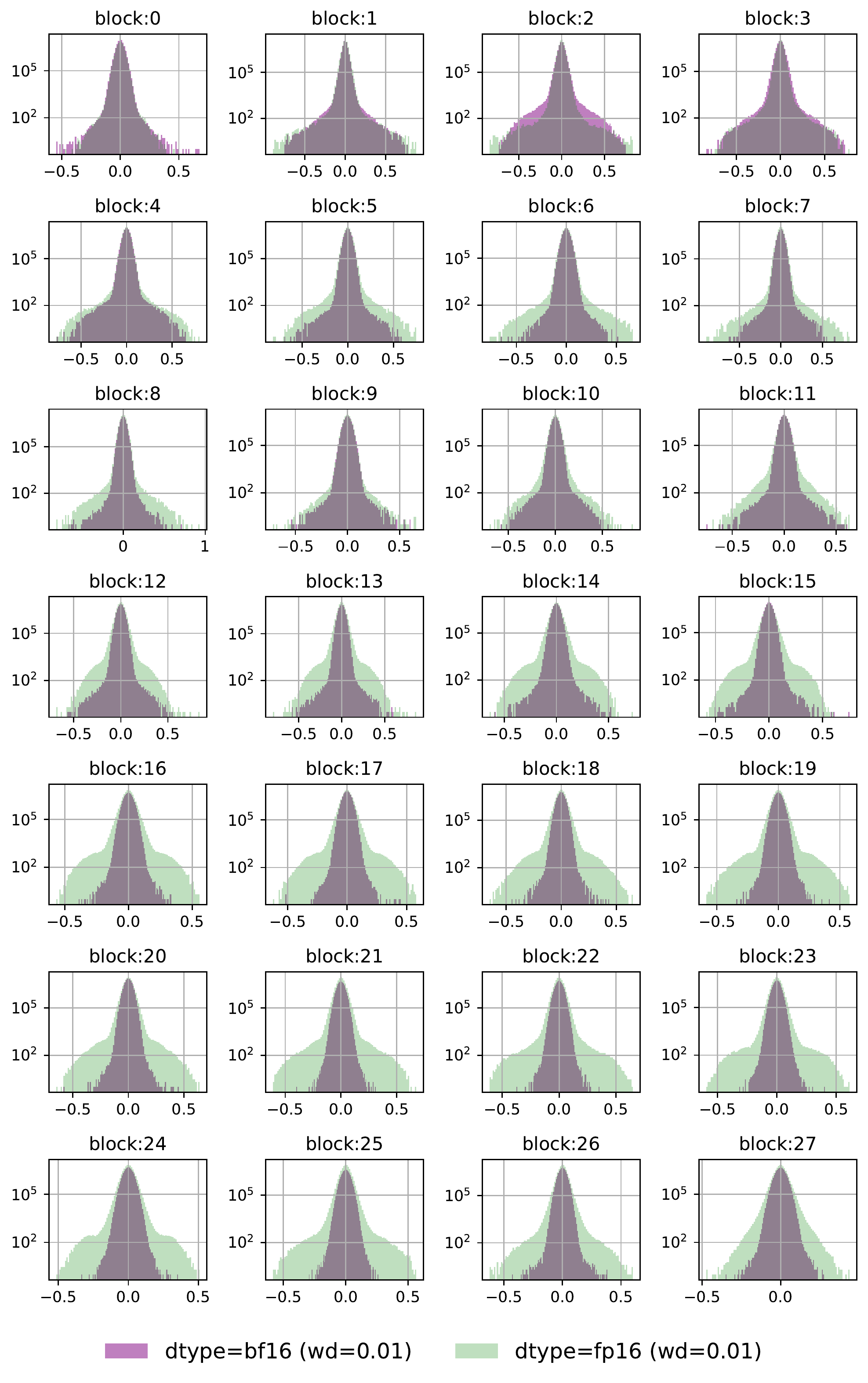}
    \caption{Weight distributions for \texttt{attn-kqv-proj} layers comparing \texttt{fp16} and \texttt{bf16} variants.}
    \label{fig:kqv_fp16_grouped}
\end{figure}
\newpage
\begin{figure}[!ht]
    \centering
    \includegraphics[height=0.91\textheight]{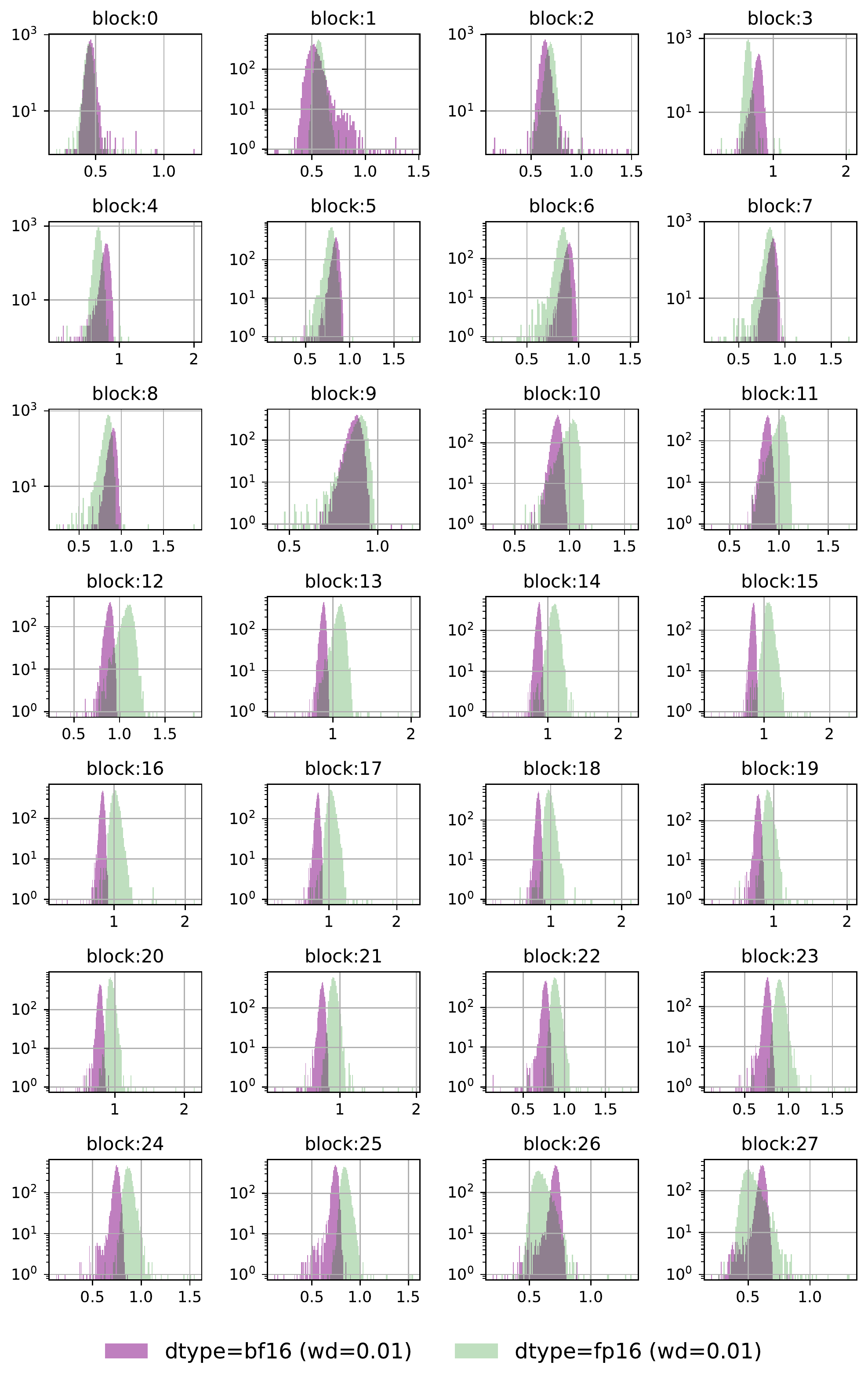}
    \caption{Gain parameter distributions of the first layernorm comparing \texttt{fp16} and \texttt{bf16} variants. }
    \label{fig:ln_fp16_grouped}
\end{figure}
\newpage
\begin{figure}[!ht]
    \centering
    \includegraphics[height=0.91\textheight]{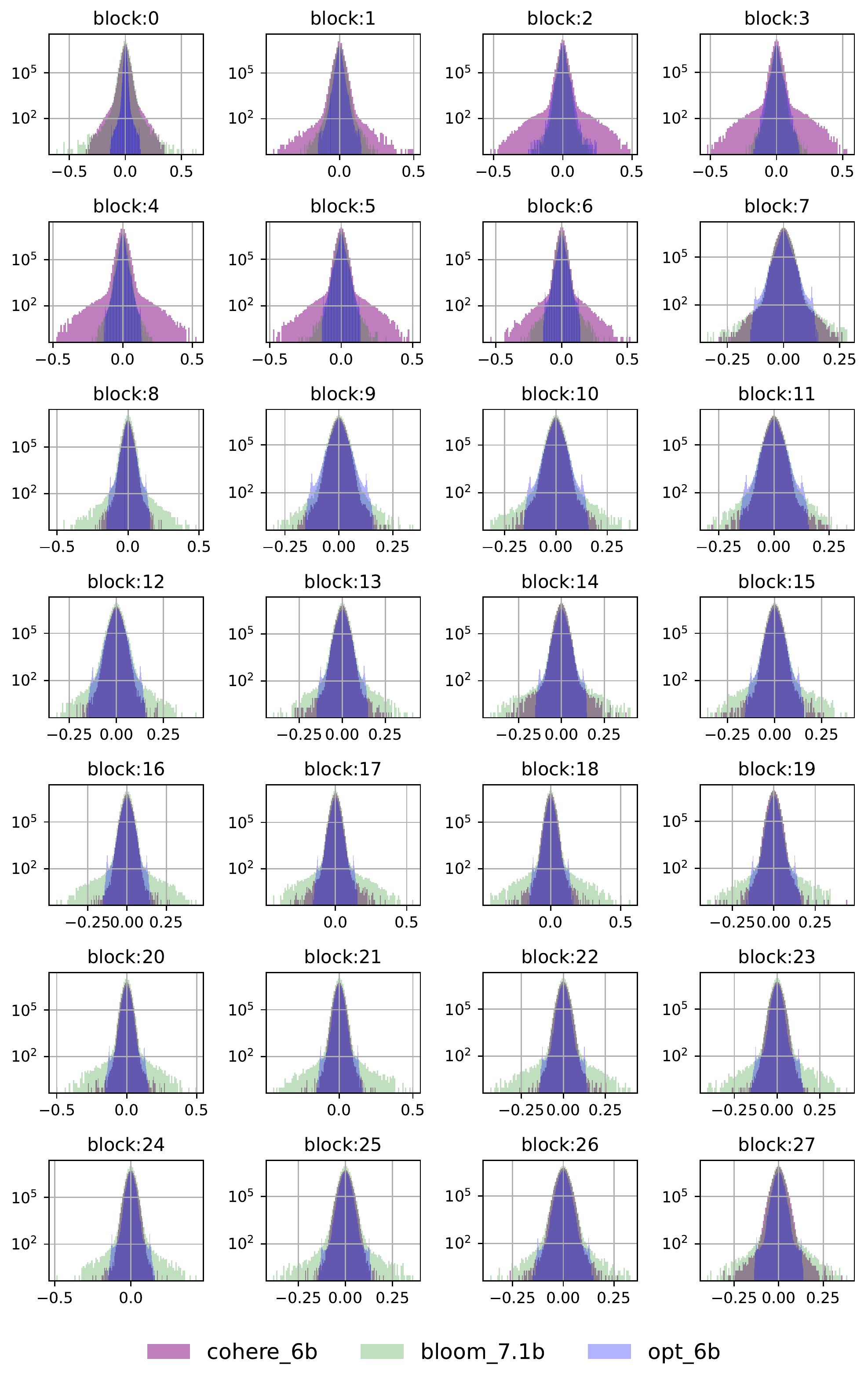}
    \caption{Weight distributions of \texttt{attn-kqv-proj} layers comparing our model against OPT-6B \& BLOOM-7.1B}
    \label{fig:kqv_prod_grouped}
\end{figure}
\newpage
\begin{figure}[!ht]
    \centering
    \includegraphics[height=0.91\textheight]{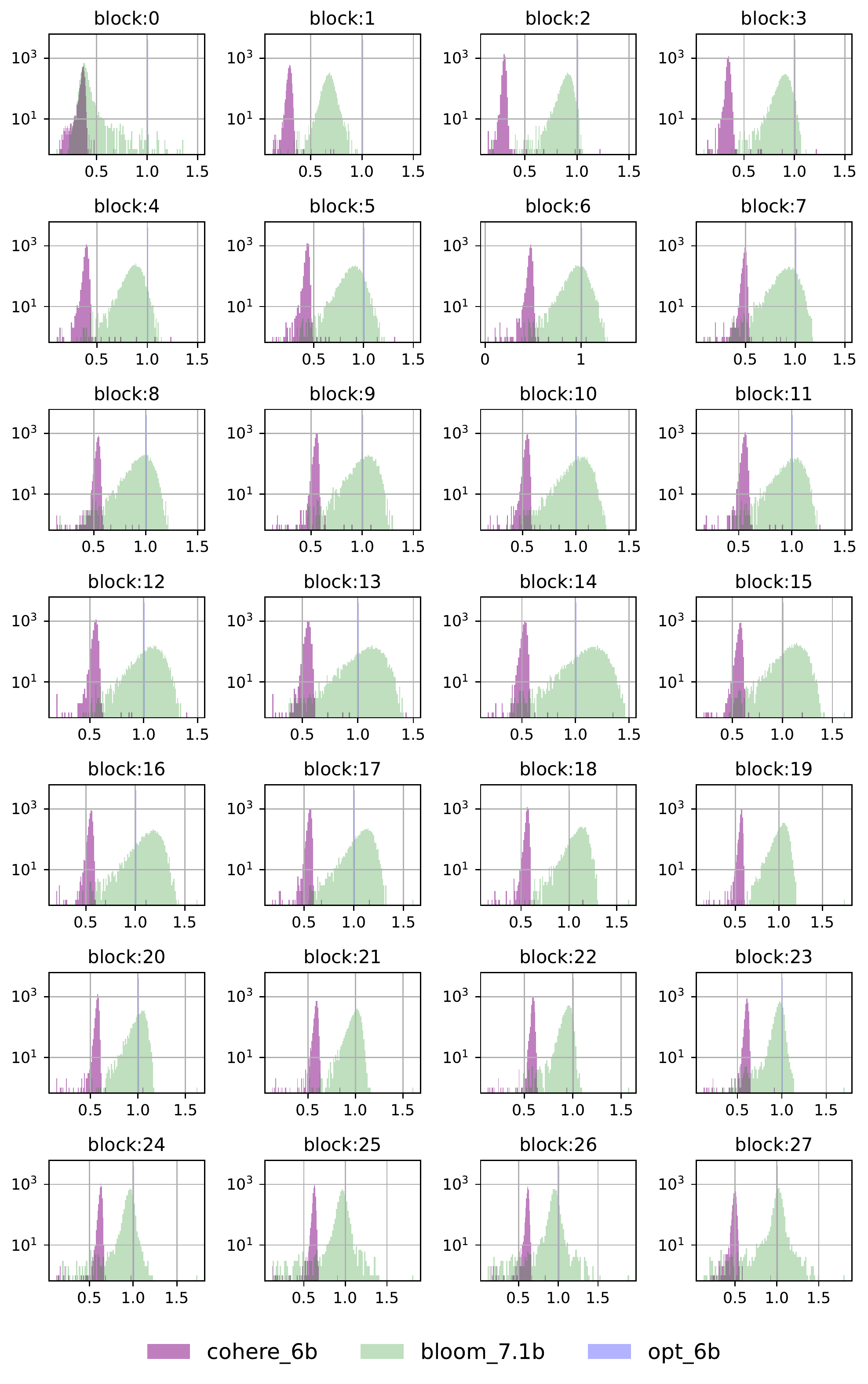}
    \caption{Gain parameter distributions of the first layernorm comparing our model against OPT-6B. Note that in OPT-6B, all  layernorm gain parameters across the are set to 1.0. }
    \label{fig:ln_prod_grouped}
\end{figure}

\newpage
\subsection{Layers Analysis}
\label{apendix:all-act-distributions}
\begin{figure}[!th]
    \centering
    \label{fig:fp16_all_layer_stats}
    \includegraphics[height=0.40\textheight]{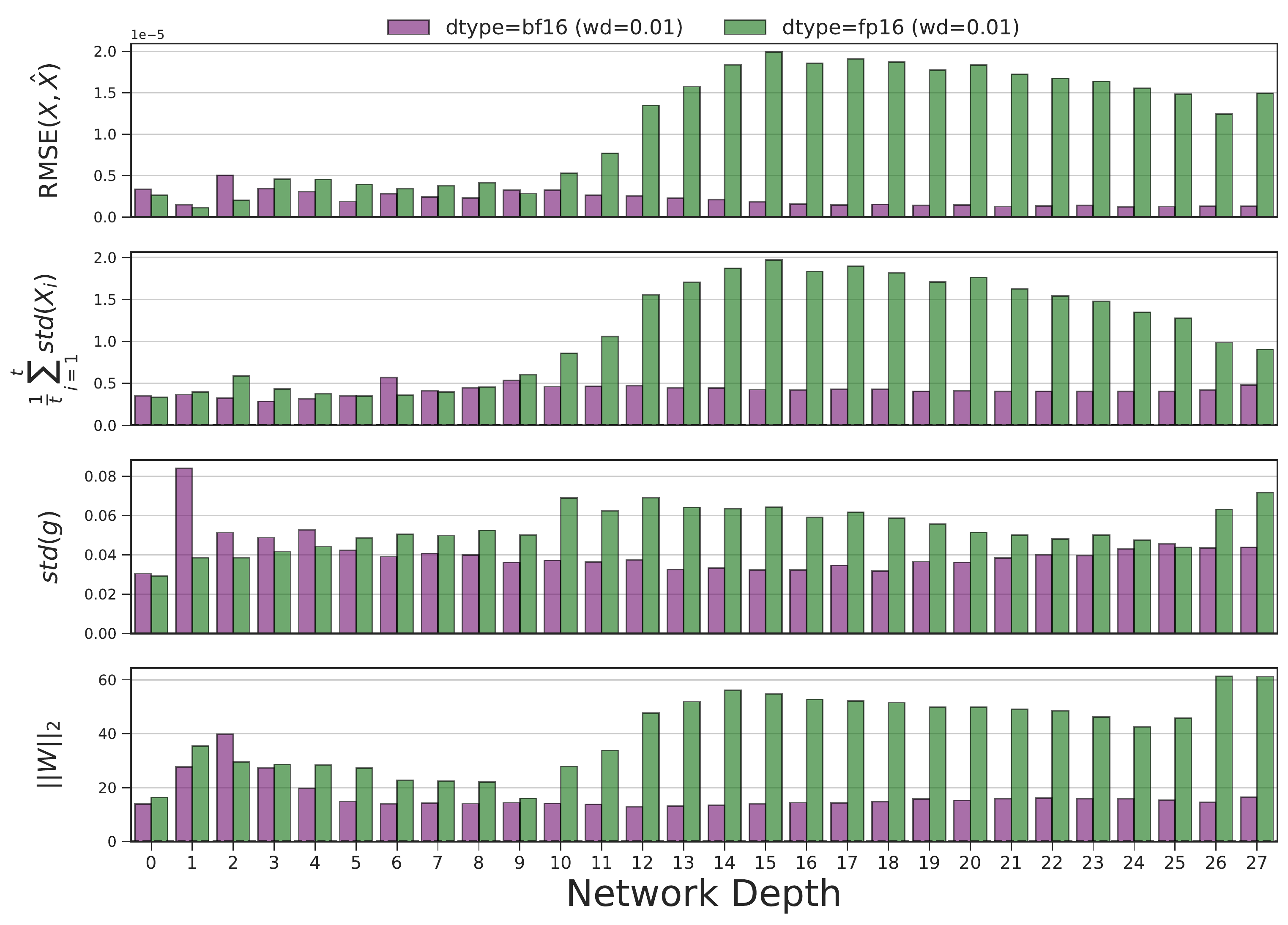}
    \caption{Input activation RMSE and token STD are higher in the \texttt{fp16} variant for most \texttt{attn-kqv-proj} layers in the network. A similar trend exists for the first layernorm gain STD and weight spectral norm.}
    \centering
    \includegraphics[height=0.40\textheight]{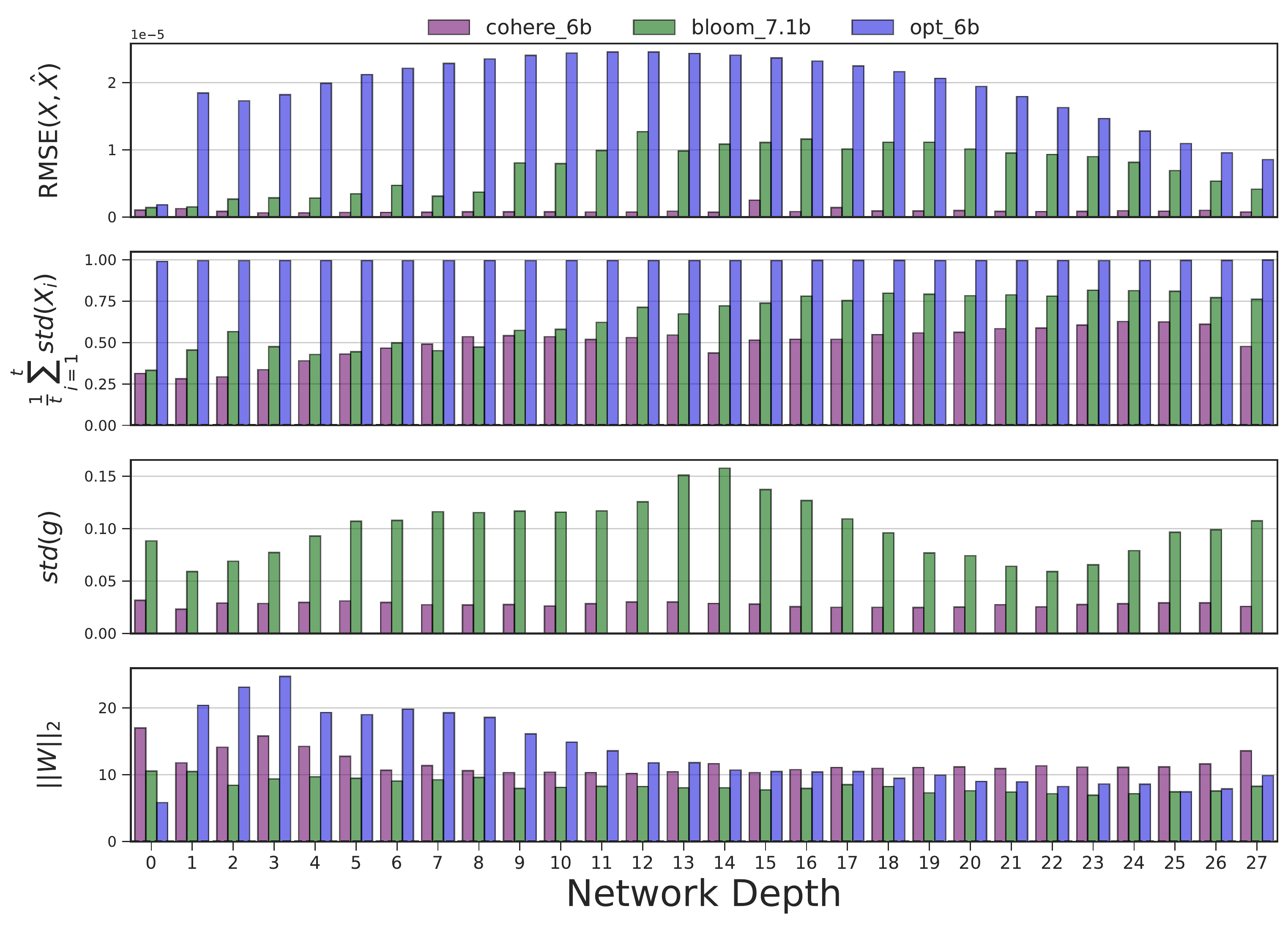}
    \caption{In line with Figure 10, the less quantization robust OPT-6B model has significantly higher RMSE($\mathbf{X}$,$\mathbf{\hat{X}}$) and average token activation STD for all \texttt{attn-kqv-proj} layers. Note that all gain parameters seem to have been hard coded to 1.0 in OPT-6B.}
    \label{fig:prod_all_layer_stats}
\end{figure}

\newpage
\subsection{Outlier analysis}  \label{apendix:outlier}
We classify a feature dimension as an outlier dimension if the \emph{same} dimension is classified as an outlier across 20 random samples from the C4 validation set \citep{raffel2020exploring}. These samples are  fixed when experimenting with different outlier definitions. As mentioned in Section \ref{sec:results}, we experimented with the definition outlined in \cite{dettmers2022gptint} where a hidden feature dimension is classified as outlier dimension 
if the activation magnitudes (denoted as $\alpha$) are greater than 6.0 ($\alpha > 6$) at more than 25\% of layers and 6\% of tokens. However, we find \emph{0} outlier dimensions across all of our 6B variants and fully trained models using this definition. As a result, following the methodology outlined in \cite{dettmers2022gptint}, we manually searched for the lowest threshold such that only one dimension is classified as outlier in our smallest (410M) fully trained model. As shown in Table \ref{tab:outliers}, even classifying outlier dimensions using the searched threshold of 4.2 resulted in most variants having \emph{0} outlier dimensions.  We further experimented with not fixing the threshold and using \textit{z-score} outlier detection i.e. classify a feature as a high-magnitude if $\alpha > C\sigma + \mu$, where $\sigma$ and $\mu$ denote sample standard deviation and mean. We were not able to establish clear trends with this method either as shown in Table \ref{tab:outliers}.

\begin{table}[thp] 
\newcommand*{\cindent}{\hspace*{0.5cm}}%
\newcommand*{\cindenttt}{\hspace*{0.80cm}}%
\newcommand*{\cindentt}{\hspace*{0.85cm}}%

\centering
\caption{Outlier statistic using different thresholding rules \textbf{Top Table:} constant threshold $\alpha > 4.2$ \textbf{Bottom Table: } adaptive threshold $\alpha > 4\sigma_{token} + \mu_{token}$ }
\begin{tabular}{p{0.30\linewidth}p{0.16\linewidth}p{0.20\linewidth}p{0.20\linewidth}}

\toprule
Variant &\#Outliers &\%Seq Affected &\%Layers Affected \\\midrule
\texttt{\{all other variants\}}  &\cindent 0 &\cindentt 0 &\cindentt 0 \\
\texttt{dtype=fp16 (wd=0.01)} & \cindent 6 & \cindenttt 68.4 & \cindenttt 65.5 \\
\midrule
\texttt{cohere\_410M} &\cindent 1 & \cindentt 25.0 & \cindentt 26.3 \\
\texttt{cohere\_6B} &\cindent0 &\cindentt0 &\cindentt0 \\

\end{tabular}
\begin{tabular}{p{0.30\linewidth}p{0.16\linewidth}p{0.20\linewidth}p{0.20\linewidth}}
\bottomrule
\\ 
\toprule
\texttt{dropout=0.1} &\cindent2 &\cindentt44.0 &\cindentt40.5 \\
\texttt{dropout=0.4} &\cindent2 &\cindentt44.9 &\cindentt42.9 \\
\texttt{dropout=0.8} &\cindent1 &\cindentt19.2 &\cindentt25 \\
\midrule
\texttt{wd=0.1 (gc=none)} &\cindent2 &\cindentt28.0 &\cindentt39.3 \\
\texttt{wd=0.01 (gc=none)} &\cindent0 &\cindentt0 &\cindentt0 \\
\texttt{wd=0.001 (gc=none)} &\cindent2 &\cindentt34.3 &\cindentt34.5 \\
\midrule
\texttt{dtype=bf16 (wd=0.1)} &\cindent2 &\cindentt33.0 &\cindentt36.9 \\
\texttt{dtype=fp16 (wd=0.1)} &\cindent2 &\cindentt41.1 &\cindentt42.9 \\
\texttt{dtype=bf16 (wd=0.01)} &\cindent0 &\cindentt0 &\cindentt0 \\
\texttt{dtype=fp16 (wd=0.01)} &\cindent8 &\cindentt66.2 &\cindentt64.3 \\
\midrule
\texttt{cohere\_410M} & \cindent55 & \cindentt88.2 & \cindentt86.2 \\
\texttt{cohere\_6B} &\cindent7 &\cindentt65.5 &\cindentt56 \\

\bottomrule
\end{tabular}
\label{tab:outliers}
\end{table}

\section{Extended Literature Review}\label{sec:ext-literature-review}

The need for compression techniques that scale to large language model settings has become increasingly urgent with larger and larger models \citep{treviso2022, yao2023comprehensive}. There has been a renewed focus on efficiency techniques \citep{gale2019,ogueji2022,ahia-etal-2021-low-resource}. Quantization as a form of model compression of large language models has become increasingly relevant as a way to minimize memory requirements and minimize compute intensity \citep{dettmers2022gptint,Guangxuan2022,frantar2022,nuQMM2022,kimwinning}. 

\textbf{Model Efficiency at Inference Time} Research in model compression mostly falls in the categories of quantization techniques \citep{Jacob_2018, 2014Courbariaux_low_precision_multiplications, Hubara2016_training_neural_networks_low_precision, 2015_gupta}, efforts to start with a network that is more compact with fewer parameters, layers or computations (architecture design) \citep{2017Howard, 2016Squeezenet, kumar17}, student networks with fewer parameters that learn from a larger teacher model (model distillation) \citep{2015hinton} and finally pruning by setting a subset of weights or filters to zero \citep{2017l0_reg, 2016learnedSparsity, Cun90optimalbrain, 1993optimalbrain, Strom97sparseconnection, Hassibi93secondorder, 2016abigail, 2017Narang,frantar2023,sanh2020movement}. Often, a combination of compression methods might be applied. For example, pruning might be combined with other efficiency-improving methods, e.g. quantization or faster search algorithms.

\textbf{Quantization Techniques}  Quantization can be used to speed up inference and relax hardware requirements, as has been shown for e.g.,  8-bit ~\citep{quinn-ballesteros-2018-pieces},  
 4-bit~\citep{aji-heafield-2020-compressing} and recently also below 3-bit quantization~\citep{nuQMM2022} of neural machine translation models. \citet{nuQMM2022} utilize non-uniform quantization to achieve high compression ratios. However, this method requires the use of specialized kernels for compressed (2-bit/4-bit) weights and floating point activations and involve finding the binary representations using expensive iterative search or QAT. Similar to \citet{nuQMM2022}, \cite{frantar2022} also demonstrate compression of model parameters to 3 or 4-bit precision allowing inference off a single A100 GPU. However, they also perform weight-only quantization limiting the speedup as the activations are kept at higher precision (FP16). 
 
 Quantization reduces the number of bits needed to represent model weights which minimizes both the memory and latency required to serve a model. 

Often, the goal is to quantize the bit representation while preserving equivalent performance. Quantization approaches can be broadly categorized into:
\begin{enumerate}
    \item \textbf{Quantization-aware training} (QAT) \citep{zafrir2019q8bert, krishnamoorthi2018quantizing} -- Quantization-aware training (QAT) involves pre-training with simulated quantization, enabling parameters to adjust to lower precision grids. This requires estimating the derivative of non-differentiable quantization operators, performing full backpropagation throughout the entire model, and training with the entire training dataset. However, this method can be computationally expensive, particularly for large language models.
    \item \textbf{Quantization-aware finetuning}  \citep{yao2022zeroquant, frantar2022, zhuo2022empirical, BRECQ2021, hubara2020improving, nagel2020up} (QAF) is a more efficient approach that utilizes a pretrained model and a small subset of training data (i.e., hundreds of samples) to optimize performance under quantization. By simulating quantization and optimizing a small range of parameters at a time, no backpropagation is needed while the quantization loss can be reduced.
    \item \textbf{One-shot post-training quantization} (PTQ) \citep{Guangxuan2022, dettmers2022gptint} unlike QAT and QAF, does not involve optimization. Instead, it directly maps data from a high precision range to a low precision range based on a hand-picked mapping function.
\end{enumerate}

Given the complexities of successfully training a large language model \citep{OPT-zhang2022,rae2021}, post-training quantization (PTQ) methods are extremely attractive as these techniques require the least modification to pretrained parameters. This is the focus of our exploration in this work.

\subsection{Introduction to Post-Training Integer Quantization Approaches}

 Below section introduces widely used quantization methods and provides context about the differences between these methods. The quantization strategy for weights and activations can be broadly classified into three categories:

\subsubsection{Weight-only Quantization} 
Weight-only quantization has proven extremely effective in making large language models accessible by enabling inference in a resource-constrained environment while maintaining the FP16 model quality \citep{llamacpp, frantar2022, Zeng2022}. Weight-only quantization provides improvements in latency due to a reduction in time taken for parameter fetching from GPU global memory, however, the actual Matrix-Matrix multiplication (GEMM) operations are carried out at higher precision in FP16 - allowing modest gains on platforms without dedicated lower-precision GEMM operations support.

\subsubsection{Weight and Activation Quantization} As large language models are scaled, progressively they become compute-bound and the improvements due to weight-only quantization stagnate. However, in this regime, using efficient kernels that leverage specialized lower-precision cores in modern GPUs to directly perform the actual Matrix-Matrix multiplication operation at lower precision enables large latency gains - due to the increased throughput of INT8 tensor cores over FP16 Tensor Cores \citep{nvidiaa100}. As this quantization technique scales the best, this is going to be our focus in this work.

To-date quantization of \emph{both} the activations and weights of very large models ($>$6.7B parameters) has proven challenging - leading to a large drop in performance.

\subsubsection{Quantization by Mixed-Precision Decomposition}

In the quantization strategies mentioned above, even though the various weights and activations might be stored in different precisions; all the computations in a single operation are carried out at the same precision (FP16 or INT8). In contrast, \texttt{LLM.int8()} \citep{dettmers2022gptint} proposes to decompose the matrix multiplication to compute a small fraction of elements at a higher precision (FP16) while the bulk of the computations is performed at low precision (INT8). 
This approach has a similar footprint to that of weight-only quantization but practical latency gains are limited or potentially worse. 
While this approach has theoretical latency benefits due to the bulk of the computation being performed at lower precision, in practice without specialized hardware \citep{9425549, 10.1145/3400302.3415679}, the lack of specialized kernels on GPUs and additional kernel calls required to ready the inputs and weights for mixed-precision computation negates the projected benefits. 
In this work we focus on exploring optimization choices which mitigate quantization trade-offs for both \textit{weight-only quantization} and the far more challenging \textit{weight and activation quantization}.

\end{document}